\documentclass[11pt,a4paper]{article}
\usepackage{times,latexsym}
\usepackage{url}
\usepackage[T1]{fontenc}

\usepackage{times}
\usepackage{latexsym}
\usepackage{graphicx}

\usepackage{color}
\usepackage{booktabs}
\usepackage{multicol}
\usepackage{graphicx}
\usepackage{xcolor}
\usepackage{afterpage}

\usepackage{caption}
\usepackage{mathtools}
\usepackage{nicematrix}
\usepackage{subcaption}
\usepackage{wrapfig}
\usepackage{colortbl}
\usepackage{bm}
\usepackage{amssymb}
\usepackage{enumitem}
\usepackage{soul}
\usepackage{tabularx}
\usepackage{float}
\usepackage{mathtools}
\usepackage[htt]{hyphenat}
\usepackage{tabu}
\usepackage{amsmath,bm}
\usepackage{array,multirow,colortbl}
\usepackage{todonotes}
\usepackage{bbm}
\usepackage{multirow}
\usepackage{float}
\usepackage{tikz}
\usepackage{svg}
\usepackage{subcaption}
\usepackage{supertabular}
\usepackage{rotating}
\usepackage{collcell}
\definecolor{GradientGreen}{HTML}{30c92a}
\definecolor{UMDYellow}{HTML}{ffc20e}
\usepackage{dsfont}
\usepackage{makecell}
\usepackage{multirow}
\usepackage{booktabs}
\usepackage{inconsolata}

\usepackage{amssymb}%
\usepackage{pifont}%

\usepackage{placeins}

   \usepackage[acceptedWithA]{tacl2021v1}

\usepackage[]{tacl2021v1}

\usepackage{xspace,mfirstuc,tabulary}

\newif\iftaclinstructions
\taclinstructionsfalse %
\iftaclinstructions
\renewcommand{\confidential}{}
\renewcommand{\anonsubtext}{(No author info supplied here, for consistency with
TACL-submission anonymization requirements)}
\newcommand{\instr}
\fi

\iftaclpubformat %

\else

\fi

\definecolor{lightgrey}{RGB}{235, 236, 237}
\definecolor{darkgrey}{RGB}{124, 124, 125}

\definecolor{lightteal}{RGB}{208,223,226}
\definecolor{teal}{RGB}{69,129,129}

\definecolor{lightorange}{RGB}{252,229,205}
\definecolor{burntorange}{RGB}{207,146,82}

\definecolor{lightpurple}{RGB}{217,210,233}
\definecolor{darkpurple}{RGB}{124,102,179}
\definecolor{coolgreen}{RGB}{73, 176, 104}
\definecolor{coolred}{RGB}{235, 125, 120}

\newcommand{\abr}[1]{\textsc{#1}}

\newcommand*{\MinNumber}{0}%
\newcommand*{\MidNumber}{100} %
\newcommand*{\MaxNumber}{200}%

\newcommand{\ApplyGradient}[1]{%
        \ifdim #1 pt > \MidNumber pt
            \pgfmathsetmacro{\PercentColor}{max(min(100.0*(#1 - \MidNumber)/(\MaxNumber-\MidNumber),100.0),0.00)} %
            \colorbox{red!\PercentColor!UMDYellow}{#1}
        \else
            \pgfmathsetmacro{\PercentColor}{max(min(100.0*(\MidNumber - #1)/(\MidNumber-\MinNumber),100.0),0.00)} %
            \colorbox{white!\PercentColor!GradientGreen}{#1}
        \fi
}

\newcolumntype{R}{>{\collectcell\ApplyGradient}c<{\endcollectcell}}

\newcommand{\bucket}{\includegraphics[height=10pt,width=10pt]{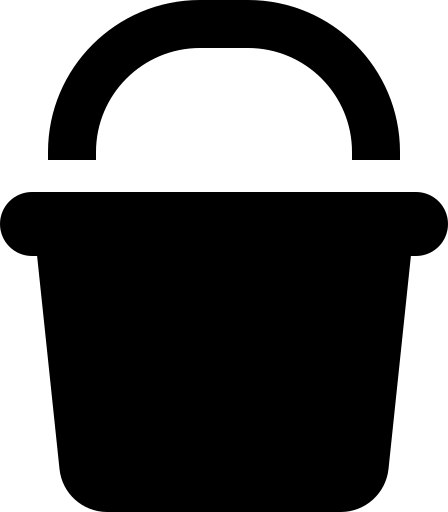}}
\newcommand{\bucketsmall}{\includegraphics[height=6pt,width=6pt]{figures/bucket-solid.png}}

\newcommand{\checkmarksmall}{\includegraphics[height=10pt,width=10pt]{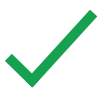}}
\newcommand{\crossmarksmall}{\includegraphics[height=10pt,width=10pt]{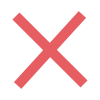}}

\newcommand{\paranlu}{\abr{ParaNlu}}
\newcommand{\roberta}{\abr{RoBERTa}}
\newcommand{\gpt}{\abr{GPT}-3}
\newcommand{\deberta}{\abr{DeBERTa}}
\newcommand{\bilstm}{\abr{BiLSTM}}

\newcommand{\pstay}{$P_C$}
\newcommand{\pstaym}{P_C}
\newcommand{\pstayc}{$\widetilde{P}_C$}
\newcommand{\anli}{$\alpha$-NLI}
\newcommand{\dnli}{$\delta$-NLI}
\newcommand{\snli}{$\delta$-SNLI}
\newcommand{\atomic}{$\delta$-ATOMIC}
\newcommand{\social}{$\delta$-SOCIAL}
\newcommand{\accoriginal}{$A_\abr{O}$}
\newcommand{\acctest}{$A_\abr{T}$}
\newcommand{\accparaphrases}{$A_{\bucketsmall}$}
\newcommand{\accparaphrasesc}{$\widetilde{A}_{\bucketsmall}$}
\newcommand{\qcpg}{\abr{QCPG}}
\newcommand{\vap}{\textbf{\abr{vap}}}
\newcommand{\pvap}{\textbf{\abr{pvap}}}
\newcommand{\premise}{\textcolor{darkpurple}{\textbf{Premise}}}
\newcommand{\hypothesis}{\textcolor{burntorange}{\textbf{Hypothesis}}}
\newcommand{\update}{\textcolor{teal}{\textbf{Update}}}

\newenvironment{packed_itemize}{
\begin{itemize}
  \setlength{\itemsep}{1pt}
  \setlength{\parskip}{0pt}
  \setlength{\parsep}{0pt}
}{\end{itemize}}

\newenvironment{tight_enumerate}{
\begin{enumerate}
  \setlength{\itemsep}{0pt}
  \setlength{\parskip}{0pt}
}{\end{enumerate}}

\title{How often are errors in natural language reasoning\\ due to paraphrastic variability?}

\author{Neha Srikanth \\
  Computer Science \\
  University of Maryland \\
  \texttt{nehasrik@umd.edu} \\\And
  Marine Carpuat \\
  Computer Science \\
  University of Maryland \\
  \texttt{marine@cs.umd.edu} \\\And
  Rachel Rudinger \\
  Computer Science \\
  University of Maryland \\
  \texttt{rudinger@umd.edu}}

\date{}

\begin{document}
\maketitle
\begin{abstract}
Large language models have been shown to behave inconsistently in response to meaning-preserving paraphrastic inputs. 
At the same time, researchers evaluate the knowledge and reasoning abilities of these models with test evaluations that do not disaggregate the effect of paraphrastic variability on performance.
We propose a metric, \pstay, for evaluating the \textit{paraphrastic consistency} of natural language reasoning models based on the probability of a model achieving the same correctness on two paraphrases of the same problem. 
We mathematically connect this metric to the proportion of a model's variance in correctness attributable to paraphrasing. 
To estimate \pstay, we collect \abr{ParaNlu}, a dataset of 7,782 human-written and validated paraphrased reasoning problems constructed on top of existing benchmark datasets for defeasible and abductive natural language inference. 
Using \abr{ParaNlu}, we measure the paraphrastic consistency of several model classes and show that consistency dramatically increases with pretraining but not fine-tuning. 
All models tested exhibited room for improvement in paraphrastic consistency.

\end{abstract}

\section{Introduction}
The NLP community has transitioned away from ``deeper'' abstract semantic representations  (e.g., FrameNet~\cite{baker1998berkeley}) towards ``shallower'' representations (e.g., Universal Dependencies~\cite{nivre2016universal}) which retain attributes of their original surface form. 
The culmination of this trend is to use natural language as a semantic representation itself for evaluating a model's reasoning ability.
This has enabled rapid advancement across a host of tasks including NLI~\cite{bowman-etal-2015-large} and QA~\cite{rajpurkar-etal-2016-squad}, with the latest generation of large language models saturating many benchmark natural language understanding datasets.
However, natural language as a meaning representation is highly ambiguous~\cite{schubert2015semantic}.
While versatile and compact, it leaves open the possibility that systems are not robust to \textit{different ways} of expressing the same meaning in natural language.

\begin{figure}[t]
\centering
\includegraphics[scale=0.67]{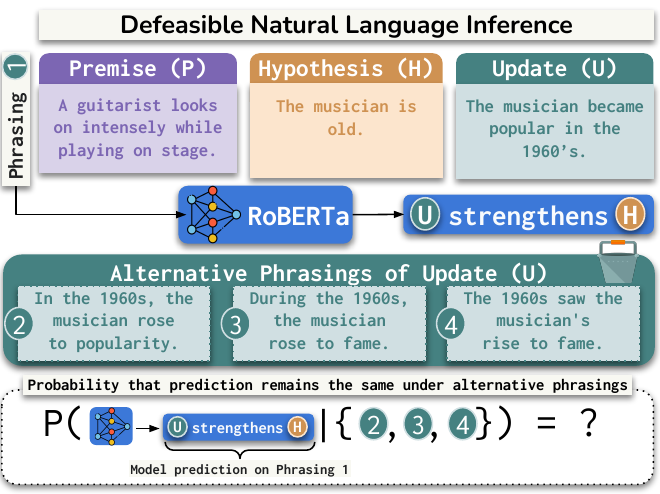}
\caption{\dnli{} instance with a set of paraphrased update sentences. We study \textbf{paraphrastic consistency}, or the probability that a model's prediction for two phrasings of the same problem match.}
\label{fig:paraphrase-example}
\end{figure}

Benchmark evaluation datasets such as SNLI~\cite{bowman-etal-2015-large} consist of a collection of reasoning problems designed to probe particular aspects of commonsense knowledge, with each example represented by a \textit{singular} linguistic expression.
When a system gets a particular example correct, it is only evidence that it was able to correctly reason \textit{for the particular phrasing used in the example}, allowing for the possibility of systems that can correctly solve one form of a reasoning problem, but not others.
Conversely, if a model gets a question wrong, how can we tell if the error was due to a failure in language understanding or a failure in reasoning? 

Consider the defeasible reasoning example in Figure \ref{fig:paraphrase-example}. A language model finetuned on the $\delta$-NLI dataset~\cite{rudinger-etal-2020-thinking} may correctly predict that the original update sentence \textit{strengthens} a human's belief in the hypothesis sentence. 
However, different linguistic expressions of that same update sentence may yield high variance in a model's predictions.
If models stay \textit{consistent} in the face of paraphrastic variability, we may conclude that correctly reasoning about one expression is indicative of an understanding of that \textit{reasoning problem}, a desirable property of teaching machines to reason entirely in natural language.

We explore the sensitivity of natural language reasoning models to paraphrasing as a way to better characterize their knowledge and reasoning ability, contextualize their performance on evaluation sets, and evaluate room for improvement on the basis of consistency.
Under the assumption that world and linguistic knowledge are separable, our study attempts to disentangle the two by generating examples that hold the required world knowledge of a reasoning problem constant while modifying its surface form, a problem formulation with connections to causality~\cite{stolfo-etal-2023-causal} and counterfactual invariance~\cite{veitch2021counterfactual, kaushik2020learning}.

To study this, we build on top of two NLU datasets---Abductive NLI~\cite{bhagavatula2019abductive} and Defeasible NLI~\cite{rudinger-etal-2020-thinking} by collecting paraphrases of reasoning problems using \textit{label-preserving} paraphrasing, a functional change of traditional paraphrasing that preserves the semantics of the \textit{core reasoning problem}.
Our dataset, \abr{ParaNlu}, contains 7,782 human-elicited and 7,295 model-elicited paraphrases across 1000 reasoning problems spanning both datasets. 
We select diverse examples to paraphrase ranging in difficulty~\cite{sakaguchi2021winogrande} and model confidence. 
Our dataset is \textit{entirely manually validated}, ensuring semantic equivalence while maximizing paraphrase diversity.

We measure \textbf{paraphrastic consistency (\pstay{})}, or the likelihood of model's prediction remaining consistent under different phrasings, in order to understand the types of surface-form changes that models are sensitive to.
We study the relationship between consistency and various data conditions and modeling paradigms, exploring factors such as data source, example difficulty, model complexity, and training dynamics.
\textit{For their given accuracy level}, we find that models still have room to improve on paraphrastic consistency.
Since no model demonstrates high accuracy \textit{and} high paraphrastic consistency, we conclude that attempts to measure their \textit{reasoning} abilities will be confounded by inconsistencies in their \textit{linguistic} abilities.

\section{Paraphrastic Consistency}
\label{sec:intuition}
\begin{figure*}[t]
\centering
\includegraphics[scale=0.75]{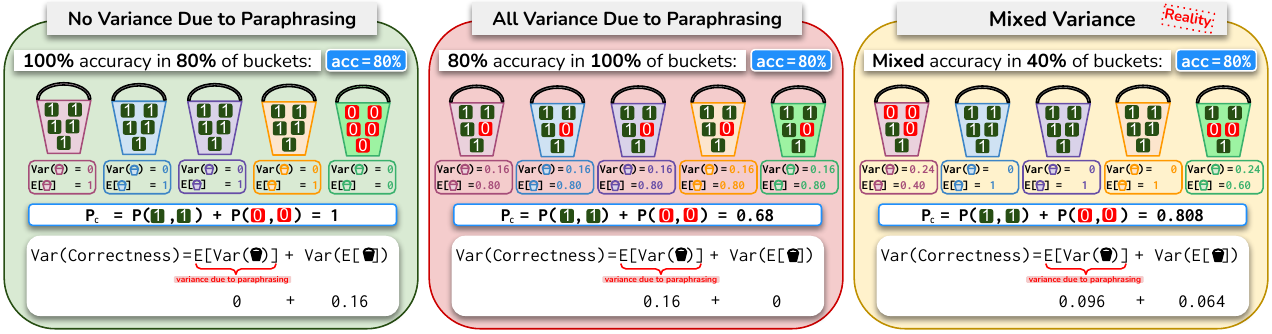}
\caption{Three scenarios, all with equivalent overall accuracy of 80\%, illustrating different distributions of variance in model predictions leading to different \pstay{} values. Buckets represent underlying commonsense reasoning problems. Numbers within buckets represent model correctness on 5 paraphrases. Most models achieve a mix of accuracy within and across buckets of paraphrased examples.}
\label{fig:bucket-cartoon}
\end{figure*}

In principle, natural language reasoning tasks like abductive NLI and defeasible NLI require the ability to linguistically \textit{decode} the meaning of the text that expresses an underlying problem, as well as the knowledge and reasoning capabilities to \textit{solve} the underlying problem.

By analogy, consider evaluating a child's understanding of the concept of addition.
Instead of simply presenting them with a mathematical expression (say, $7+7$), we write a \textit{word problem} that can be answered by (1) understanding the situation in natural language, (2) recognizing that the answer corresponds to the mathematical reasoning problem $7+7$, and finally, (3) solving $7+7$.
If the child answers incorrectly, we must figure out whether they did not understand the goal of the word problem or were not able to perform the arithmetic in order to evaluate their mathematical reasoning ability.

For models tasked with natural language reasoning problems, teasing apart these two failure modes (namely, deficiencies in language understanding versus deficiencies in knowledge or reasoning) requires more than reporting test set \textit{accuracy}.
The design of natural language reasoning test sets does not facilitate this type of analysis: if a test set contains 100 different natural language reasoning problems, and a model correctly answers 80\% of them, which failure mode should we attribute the 20\% of errors to?

For practitioners, it is useful to characterize performance by measuring \textit{paraphrastic consistency} alongside accuracy: how likely is it that a model's prediction for a natural language reasoning problem will remain the same given a \textit{different phrasing} of the problem?
We collect a dataset that changes the \textit{language} of NLI examples while maintaining the \textit{underlying logic} of the problem to tease the two apart.
For a test example, $x$, we collect a set of paraphrases $\{x'_1,x'_2,...\}$, which we call a \textit{bucket} (\bucket).
After collecting many such buckets, we can directly estimate the probability that a model's prediction for any two paraphrases belonging to the same bucket will be the same.

\subsection{Measuring Paraphrastic Consistency}

When authoring a test example for a natural language reasoning task, a crowdworker has many linguistic choices for expressing the underlying reasoning problem. 
If the purpose of the resulting test set is to evaluate a model's ability to perform the underlying reasoning task, then ideally the crowdworker's choice of phrasing would have no effect on the model's performance. 
In practice, however, it is known that language models exhibit some degree of sensitivity to paraphrastic variation ~\cite{verma-etal-2023-evaluating, jiang-etal-qa-calibration}. 
To quantify this effect, we pose the following counterfactual question: \textbf{If a given test question had been written differently, what is the probability that a model would still receive the same credit for its prediction on the paraphrased question?}

Quantitatively, we introduce a metric of \textit{paraphrastic consistency}, \pstay, defined as the probability that a model's predictions for two paraphrases of the same problem, $x'_i$ and $x'_j$, are either both correct or both incorrect (provided the ground truth labels of $x'_i$ and $x'_j$ match). To formalize \pstay, we define the following terms:
\begin{packed_itemize}
  \item $R_{\bucketsmall}$: a discrete random binary variable indicating whether a model's prediction of a paraphrased reasoning problem, $x'$, is correctly predicted (1) or incorrectly predicted (0). 
  \item $\theta_{\bucketsmall}$: $\mathds{E}[R_{\bucketsmall}]$, or the average correctness (i.e., accuracy) of the paraphrased problems ($x'_1, x'_2, ...$) in a particular bucket.
  \item $A$: Overall accuracy of model $M$ over a set of natural language reasoning problems. This is equivalent to $\mathds{E}[\theta_{\bucketsmall}]$ across all buckets.
\end{packed_itemize}

\noindent For a binary classification task, where $y \in {0,1}$ denotes the gold label, we then define \pstay{} as: 
\vspace{-0.5em}
\begin{multline}
    \pstaym = \underbrace{P(M(x_i)= y, M(x_j) = y)}_{\text{prob. of both predictions correct}} + \\
    \underbrace{P(M(x_i) \neq y, M(x_j) \neq y)}_{\text{prob. of both predictions incorrect}}
\end{multline}

\noindent \pstay~can be estimated from the accuracies of paraphrase buckets, $\theta_{\bucketsmall}$, as follows:
\vspace{-0.5em}
\begin{equation}
    \pstaym = \mathds{E}[\theta_{\bucketsmall}^2] + \mathds{E}[(1-\theta_{\bucketsmall})^2]
\label{eq:pstay}
\end{equation}

\noindent Informally, the \textit{paraphrastic consistency} of a natural language reasoning model is its ability (or lack thereof) to make the same predictions on paraphrased inputs that, in principle, should yield the same answer.
We note that the metric \pstay~\textit{cannot be computed from a standard test set containing only one phrasing per test example}; to estimate \pstay~we collect paraphrases of test examples, as described in \S\ref{sec:paranlu}.

Figure~\ref{fig:bucket-cartoon} illustrates the computation of \pstay{} for different patterns of model behavior.
If $\pstaym=1$, the model is either entirely correct or incorrect on all paraphrases of the same problem, as in the left-most panel of Figure~\ref{fig:bucket-cartoon} where each bucket contains only 1's or only 0's. 
In this case, no errors can be attributed to paraphrastic variance. 
The minimum \pstay~occurs when every paraphrase bucket has the same accuracy, as shown in the middle panel; in this case \textit{all} errors are likely due to paraphrastic variability. 
In practice, \pstay~lies between these two extremes and \textit{some}, but not all, errors are due to paraphrasing, as in the right-most panel.
Modern NLP evaluation sets usually consist of a collection of independent reasoning problems each represented by singular natural language expression, and practitioners often make claims about the reasoning capabilities or world knowledge given a model's accuracy on such evaluation sets.
However, as depicted in Figure~\ref{fig:bucket-cartoon}, accuracy presents an incomplete picture of performance: in all three scenarios, the overall accuracy remain 80\%, but only the first scenario, in which the model makes equivalent predictions given many alternate phrasings of a reasoning problem, results in a high \pstay{}.

\paragraph{\pstay~in Practice.} \pstay{} can be interpreted as the probability that given two phrasings of the same reasoning problem, the model's two predictions are either both correct or both incorrect.
This summary metric allows us to capture the \textit{reliability} of a model's prediction: how confident can we be that a model's prediction would remain correct (or incorrect) if it had been phrased differently?
While \pstay{} is lower-bounded by a function of accuracy, we design it as a metric \textit{complementary} to accuracy in order to better characterize model performance, diagnose modeling errors, and benchmark the linguistic reasoning capabilities.
For example, when two models achieve similar accuracies, computing their respective paraphrastic consistencies may help as an additional point of comparison. 
Just as desired model accuracy may be dependent on the application in which it is deployed, different settings may mandate varying appropriate \pstay{} values and practitioners must decide on tolerable thresholds based on their use case.

\subsection{Proportion of Variance Attributable to Paraphrasing}
Decomposing the \textit{total variance in correctness} across all paraphrased examples in all buckets gives us a clearer picture of the two failure modes described in the opening of \S\ref{sec:intuition}.
Using the law of total variance \cite{weiss2006course}, we decompose the total variance of the correctness across all \textit{paraphrased} examples (the variance of $R$ for all paraphrased problems in all buckets) into two terms: the average variance of correctness within a bucket, $\mathds{E}[\mathrm{Var}(R_{\bucketsmall})]$, and the variance in mean correctness (accuracy) across buckets, $\mathrm{Var}(\mathds{E}[(R_{\bucketsmall})])$, or simply, $\mathrm{Var}(\theta_{\bucketsmall})$.
\vspace{-0.05em}
\begin{equation}
    \mathrm{Var}(R) = \underbrace{\mathds{E}[\mathrm{Var}(R_{\bucketsmall})]}_{\text{variance from  paraphrasing}} + \mathrm{Var}(\theta_{\bucketsmall})
\label{eq:total-variance-law}
\end{equation}

\noindent This breakdown allows us to better identify the \textit{source} of the variance in correctness.
The first term, commonly known as \textit{unexplained variance}, measures \underline{variance} \underline{attributable} to \underline{paraphrasing} (henceforth denoted as \vap{}) within a bucket.
The second, commonly known as \textit{explained variance}, represents variance \textit{across buckets} due to inherent differences in latent characteristics (e.g. difficulty) of different problems.
If paraphrasing has no effect, the variance of correctness in each bucket ($\mathrm{Var}(R_{\bucketsmall})$) is zero and consequently, the \vap{} should also be zero. 
Most evaluation paradigms cannot directly measure \vap{}, since during data collection, multiple surface forms of the same reasoning problem are not collected.
We replicate the conditions under which original examples were produced to simulate different linguistic expressions annotators \textit{could} have chosen.

Mathematical manipulation yields: 
\vspace{-0.5em}
\begin{multline}
    \pstaym = 1 - \underbrace{P(M(x_i)=y, M(x_j)\neq y)}_{\text{prediction flips to incorrect}} \\ -
    \underbrace{P(M(x_i)\neq y, M(x_j)= y)}_{\text{prediction flips to correct}}
\end{multline}
\vspace{-1em}
\begin{multline}
    \pstaym = 1 - 
    \underbrace{\mathds{E}[\theta_{\bucketsmall} \cdot (1-\theta_{\bucketsmall})]}_{\text{prediction flips to incorrect}} \\ - 
    \underbrace{\mathds{E}[(1-\theta_{\bucketsmall}) \cdot \theta_{\bucketsmall}]}_{\text{prediction flips to correct}}
\end{multline}
\vspace{-1em}
\begin{align}
    \pstaym &= 1 - 2 \cdot \mathds{E}[\theta_{\bucketsmall} \cdot (1-\theta_{\bucketsmall})] \\
    &= 1 - 2 \cdot \underbrace{\mathds{E}[\mathrm{Var}(R_{\bucketsmall})]}_{\vap{}}
    \label{eq:uv-pstay}
\end{align}

\noindent Although the primary metric we report throughout the paper is \pstay, Equation~\ref{eq:uv-pstay} highlights the direct negative relationship between \pstay~and variance attributable to paraphrasing (\vap{}).

Related to \vap{} is the \textit{proportion} of total variance attributable to paraphrasing (\pvap{}), which is simply \vap{} divided by the total variance:
\vspace{-0.5em}
\begin{align}
    \mathrm{PVAP} = \mathrm{VAP} / \mathrm{Var}(R)
    \label{eq:pvap}
\end{align}

\paragraph{Model Confidence and \pstay{}.} We choose to characterize paraphrastic consistency via variance in \textit{correctness} instead of variance in \textit{model confidence}. 
Confidence represents a model's \textit{overall} estimate in the correct answer, and thus conflates confidence in linguistic understanding and problem solving.
A low confidence in the \textit{correct label} may indicate that the model understood what the problem was asking, but was unable to reach the answer (or vice versa).
Models trained to optimize for \textit{accuracy} are not calibrated to explicitly encode confidence in linguistic decoding or problem-solving ability.

\section{Reasoning Tasks}
\begin{table*}[ht!]
\centering
\resizebox{2.08\columnwidth}{!}{
    \begin{tabular}{p{0.35\linewidth}p{0.35\linewidth}p{0.35\linewidth}p{0.35\linewidth}} %
    \toprule
    \multicolumn{1}{c}{\textbf{\textbf{Context ($C$)}}} & \multicolumn{1}{c}{\textbf{Original Target ($T$)}} & \multicolumn{1}{c}{\textbf{Admissible Paraphrase}} & \multicolumn{1}{c}{\textbf{Inadmissible Paraphrase}} \\ 
    \midrule
    \colorbox{lightpurple}{\textcolor{darkpurple}{\textbf{O1:}}} My cell phone lanyard broke on Wednesday \newline 
    \colorbox{lightorange}{\textcolor{burntorange}{\textbf{O2:}}} My wife was able to repair the lanyard so I did not have to buy one. &
    \colorbox{lightteal}{\textcolor{teal}{\textbf{$h^+$}:}} It looked like it was ripped or torn. \newline \colorbox{lightteal}{\textcolor{teal}{\textbf{$h^-$}:}} It didn't look like it was ripped or torn. &
    \colorbox{lightteal}{\textcolor{teal}{\textbf{$h^+$}:}} It seemed to be split and frayed. \newline \colorbox{lightteal}{\textcolor{teal}{\textbf{$h^-$}:}} It did not appear to be damaged. &
    \colorbox{lightteal}{\textcolor{teal}{\textbf{$h^+$}:}} The screen had wear and tear. \newline \colorbox{lightteal}{\textcolor{teal}{\textbf{$h^-$:}}} It looked like it was in good shape, but apparently it wasn't.\\
    \midrule
    \colorbox{lightpurple}{\textcolor{darkpurple}{\textbf{O1:}}} Mike was graduating high school later in the day. \newline 
    \colorbox{lightorange}{\textcolor{burntorange}{\textbf{O2:}}} When he got back home, Mike regretted skipping graduation. &
    \colorbox{lightteal}{\textcolor{teal}{\textbf{$h^+$:}}} Mike's parents said they couldn't go, but they did. \newline 
    \colorbox{lightteal}{\textcolor{teal}{\textbf{$h^-$:}}} Mikes parents didn't show up. &
    \colorbox{lightteal}{\textcolor{teal}{\textbf{$h^+$:}}} Despite Mike's parents' claims that they couldn't, they went. \newline 
    \colorbox{lightteal}{\textcolor{teal}{\textbf{$h^-$:}}} Mike's parents were absent from the graduation ceremony. &
    \colorbox{lightteal}{\textcolor{teal}{\textbf{$h^+$:}}} His parents mentioned to him they couldn’t go. \newline 
    \colorbox{lightteal}{\textcolor{teal}{\textbf{$h^-$:}}} Nobody showed up to see Mike's parents.\\
    \toprule
    \colorbox{lightpurple}{\textcolor{darkpurple}{\textbf{P:}}} PersonX stops eating fast food. \newline
    \colorbox{lightorange}{\textcolor{burntorange}{\textbf{H:}}} PersonX is seen as better. &
    \colorbox{lightteal}{\textcolor{teal}{\textbf{$U:$}}} PersonX's high school friends think this is strange. &
    \colorbox{lightteal}{\textcolor{teal}{\textbf{$U:$}}} This was considered weird by PersonX’s friends from high school. &
    \colorbox{lightteal}{\textcolor{teal}{\textbf{$U:$}}} PersonX's alumni from high school find this to be peculiar.\\
    \toprule
    \colorbox{lightpurple}{\textcolor{darkpurple}{\textbf{P:}}} Two men walking down the sidewalk carrying skateboards. \newline
    \colorbox{lightorange}{\textcolor{burntorange}{\textbf{H:}}} Two people are walking to the skate park. &
    \colorbox{lightteal}{\textcolor{teal}{\textbf{$U:$}}} Both men are adjusting their wheels as they walk. &
    \colorbox{lightteal}{\textcolor{teal}{\textbf{$U:$}}} As they stride along, both men are changing the settings on their wheels. &
    \colorbox{lightteal}{\textcolor{teal}{\textbf{$U:$}}} While walking the men re-wax their boards. \\
    \toprule
    \colorbox{lightorange}{\textcolor{burntorange}{\textbf{H:}}} It's okay to cancel an job interview. &
    \colorbox{lightteal}{\textcolor{teal}{\textbf{$U:$}}} You are too lazy to get out of bed. &
    \colorbox{lightteal}{\textcolor{teal}{\textbf{$U:$}}} You lack the motivation to leave the comfort of your bed. &
    \colorbox{lightteal}{\textcolor{teal}{\textbf{$U:$}}} You’re feeling sick and can’t make yourself rise from bed. \\
    \bottomrule
    \end{tabular}
}
\caption{Examples of paraphrased abductive NLI (rows 1 and 2) and defeasible NLI (rows 3 -- 5)  problems in \abr{ParaNlu}. We show paraphrases written by crowdworkers that are admissible under our definition of label-preserving paraphrasing, and rejected paraphrases that did not meet the label-preserving criteria.}
\label{table:task-invariant-paraphrasing-examples}
\end{table*}

We study paraphrastic consistency across two commonsense reasoning tasks: defeasible reasoning (\S\ref{sec:defeasible}) and abductive reasoning (\S\ref{sec:abductive}). 

\subsection{Defeasible Reasoning}
\label{sec:defeasible}
Defeasible reasoning is a mode of reasoning in which inferences or conclusions may be altered or withdrawn in light of new evidence~\cite{reiter1980logic}. 
For example, given the context \textit{``A group of people perform on a stage''}, the natural conclusion \textit{``A group performs for an audience''} may be weakened upon learning that the group is at a rehearsal.

To study defeasible reasoning in language models, \citet{rudinger-etal-2020-thinking} introduce the task of defeasible natural language inference (NLI).
Traditionally, NLI involves determining whether a \textbf{premise} $P$ entails, contradicts or is neutral in relation to a \textbf{hypothesis} $H$~\cite{giampiccolo-etal-2007-third}.
When the premise $P$ and hypothesis $H$ are neutral in relation to one another, defeasible NLI studies whether a third \textbf{update} ($U$) sentence strengthens or weakens $H$.
Namely, a human may determine $H$ more likely to be true when $U$ is a strengthener, and less likely to be true when $U$ is a weakener.

\begingroup
\addtolength\leftmargini{-0.2in}
\begin{quote}
\small
    \colorbox{lightpurple}{\textcolor{darkpurple}{\textbf{Premise:}}} \hspace{0.1em} A woman in shorts throwing a bowling ball down a bowling alley. \newline
    \colorbox{lightorange}{\textcolor{burntorange}{\textbf{Hypothesis:}}} \hspace{0.1em} A woman is getting a strike! \newline
    \colorbox{lightteal}{\textcolor{teal}{\textbf{Update:}}} \hspace{0.1em} The bowling ball falls in the gutter first. 
    \colorbox{lightgrey}{\textcolor{darkgrey}{\textbf{Update Type:}}} \hspace{0.1em} Weakener
    
\end{quote}
\endgroup

\noindent \citet{rudinger-etal-2020-thinking} also introduce \dnli{}, a dataset that extends three existing natural language datasets: SNLI~\cite{bowman-etal-2015-large}, SOCIAL-CHEM-101~\cite{forbes-etal-2020-social}, and ATOMIC~\cite{sap2019atomic}.
For each premise-hypothesis pair (or just hypothesis, in the case of \social{}), crowdworkers write multiple strengthening and weakening updates, ensuring a balance.
Defeasible NLI is a binary classification task that involves predicting whether the update sentence is a strengthener or a weakener (e.g. the original update in Row 5 of Table~\ref{table:task-invariant-paraphrasing-examples} is a weakener).

Adopting the terminology from \citet{srikanth-rudinger-2022-partial}, we distinguish between \textit{context} parts of examples, and \textit{target} parts of examples, with our study of consistency concerning \textit{target} portions of examples.
For \dnli{}, we consider the premise $P$ and hypothesis $H$ as \textit{context} sentences, and the update $U$ sentence as the target (Table~\ref{table:task-invariant-paraphrasing-examples}).

\subsection{Abductive Reasoning}
\label{sec:abductive}
Abduction is inference to the most likely explanation~\cite{peirce1974collected}. Such inferences are hypotheses that can best fit one or more incomplete observations.
\begingroup
\addtolength\leftmargini{-0.2in}
\begin{quote}
\small
    \colorbox{lightpurple}{\textcolor{darkpurple}{\textbf{Observation 1:}}} \hspace{0.1em} George decided to buy a TV. \newline
    \colorbox{lightorange}{\textcolor{burntorange}{\textbf{Observation 2:}}} \hspace{0.1em} It turned out just as he'd hoped. \newline
    \colorbox{lightteal}{\textcolor{teal}{\textbf{Hypothesis $-$:}}} \hspace{0.1em} The TV had a cracked screen.
    \colorbox{lightteal}{\textcolor{teal}{\textbf{Hypothesis $+$:}}} \hspace{0.1em} Upon taking it home and unpacking it, he placed it where he wanted it.
\end{quote}
\endgroup
\noindent Given the example above, any human would most likely infer the second hypothesis over the first to explain the two observations. 
\citet{bhagavatula2019abductive} introduce abductive NLI, an abductive reasoning task formulated binary classification. \anli{} examples consist of two observations $O_1$ and $O_2$ (where $O_2$ occurs some point in time after $O_1$), $h^{+}$, a plausible hypothesis, and $h^{-}$, an implausible hypothesis. 
Given both observations, the task is to determine which hypothesis is more plausible. 
We treat $O_1$ and $O_2$ as context ($C$) and $h^{+}$ and $h^{-}$ as the target ($T$) portion of the example.

\section{Constructing \abr{ParaNlu}}
\label{sec:paranlu}
We study paraphrastic consistency by paraphrasing the target  portions ($T$) of examples from  \anli{} and all three data sources in \dnli{} (\snli{}, \atomic{}, and \social{}).
For an original example $x$, we collect a \textit{bucket} of paraphrased examples $x_i$ such that the context portions $C$ and gold label $l$ remain identical, while the target portions $T$ are rewritten as \textit{label-preserving paraphrases}.
This allows us to modify the surface form of the example while retaining the underlying commonsense reasoning problem.

\paragraph{Label-Preserving Paraphrases.} We construct quasi-paraphrases~\cite{bhagat2013paraphrase} of \textit{target} sentences in reasoning problems by loosening the requirement of semantic equivalence.
Given a natural language reasoning task $L$, a textual instance $x$ of task $L$ with label $\ell_L(x)$, we say that $x'$ is a \textit{label-preserving} paraphrase of $x$ if:
\vspace{-0.5em}
\begin{tight_enumerate}
    \item $\ell_L(x) = \ell_L(x')$
    \item $x'$ does not contradict any context ($C$) in $x$.
    \item $x'$ remains consistent with the situation evoked by $x$.
\end{tight_enumerate}

\noindent Label-preserving paraphrases are \textit{functionally} equivalent: target sentences (hypotheses in \anli{} and updates in \dnli{}) may introduce small bits of information as long as the same scenario is plausibly described. 
Consider the following \dnli{} example: \colorbox{lightpurple}{\textcolor{darkpurple}{\textbf{P:}}} \textit{A man stands in front of a cashier and kiosk at a grocery store} 
\colorbox{lightorange}{\textcolor{burntorange}{\textbf{H:}}} \textit{He is smiling}
\colorbox{lightteal}{\textcolor{teal}{\textbf{U:}}} \textit{``The man got a \underline{discount}.''}
While the sentence \textit{``The man saved \underline{10\% with a coupon}''} is not semantically equivalent to the strengthening update sentence, it is a valid, label-preserving paraphrase since it retains the logic of the problem and the label.
Table~\ref{table:task-invariant-paraphrasing-examples} shows examples of admissible and inadmissible paraphrases under our definition.
Label-preserving paraphrases represent alternative, but equivalent expressions annotators \textit{could} have chosen when writing the original problem that employ similar world knowledge.
This is different from \textit{label-altering} edits proposed by ~\citet{gardner-etal-2020-evaluating}, where minimal human edits that shift the target label were used to create examples for measuring linguistic robustness.  

We describe our example selection process for annotation (\S\ref{sec:example-selection}) and crowdsourcing protocol for collecting paraphrases and summary statistics of our dataset (\S\ref{sec:mturk}).

\begin{table}[t!]
\centering
\small
\setlength\tabcolsep{1.3pt}
\begin{tabular}{lccc} 
\toprule
 & \begin{tabular}[c]{@{}c@{}}\textbf{\# }\\\textbf{original}\end{tabular} & \begin{tabular}[c]{@{}c@{}}\textbf{\# }\\\textbf{paraphrases}\end{tabular} & \begin{tabular}[c]{@{}c@{}}\textbf{mean \#}\\\textbf{paraphrases/ex}\end{tabular} \\ 
\midrule
\textbf{$\alpha$-NLI} & 250 & 2098 & 8.4 $\pm$ 1.2 \\
\textbf{$\delta$-SNLI} & 250 & 1980 & 7.9 $\pm$ 1.4 \\
\textbf{$\delta$-ATOMIC} & 250 & 1869 & 7.5 $\pm$ 1.6 \\
\textbf{$\delta$-SOCIAL} & 250 & 1835 & 7.3 $\pm$ 1.8 \\
\bottomrule
\end{tabular}
\caption{We sample 250 problems from $\alpha$-NLI and $\delta$-NLI datasets. Total number and mean number of paraphrases depended on validation.}
\label{tab:paraphrase-dataset-sizes}
\end{table}

\subsection{Original Example Selection}
\label{sec:example-selection}
We adopt a stratified sampling strategy to obtain diverse examples for annotation that vary in difficulty.
To obtain such examples, we leverage \abr{AFLite}~\cite{sakaguchi2021winogrande}, an adversarial filtering~\cite{zellers2018swag} algorithm designed to partition datasets based on difficulty using pre-computed dense embeddings of examples fed into an ensemble of $n$ logistic regression classifiers.
At each iteration of \abr{AFLite}, members of the ensemble are trained on random partitions of $m$ examples and evaluated on the remaining validation examples. 
Each validation example is assigned a score, computed by the proportion of correct predictions. 
The top-$k$ examples with scores above a threshold $\tau$ are subsequently added to the easy partition of the dataset and filtered out, and the process repeats until less than $k$ instances are removed in a particular iteration.
Including both types of examples, easy and hard, ensures that \paranlu{} can support analysis that investigates whether models are inconsistent only on certain \textit{classes} of examples (e.g. those filtered out due to lexical artifacts).  

\paragraph{Defeasible NLI.}
We use the train splits of \snli{}, \atomic{}, and \social{} to source examples for annotation, since they are large enough to meaningfully partition using \abr{AFLite}. 
We partition each train set into 3 sections: (1) examples to finetune the \roberta-base models used to embed examples ($\roberta_{\text{embed}}$), (2) training examples for models used for consistency analysis ($\roberta_{\text{analysis}}$), and (3) a pool from which examples are sampled for annotation. 
We partition at the premise-hypothesis ($P$-$H$) level to avoid leakage, since multiple examples may contain the same $P$-$H$ pair, but different updates ($U$).

For each dataset, we pre-compute example embeddings with the $\roberta_{\text{embed}}$ model, and run \abr{aflite} with the  $n=64$ linear classifiers, $\tau=0.75$, $k=500$, and $m=5000$.\footnote{We keep all hyperparameters unchanged from \citet{sakaguchi2021winogrande} with the exception of $m$ which we reduce from $10000$ to $5000$ to account for the smaller training set partitions, as in \citet{herlihy-rudinger-2021-mednli}.}
Examples in the annotation pool with $\tau \leq 0.75$ are added to the \textit{easy} subset, and the those with $\tau > 0.75$ are labeled as \textit{difficult}.
We then finetune a separate \roberta-large model on $\roberta_{\text{analysis}}$ and use it to obtain predictions on examples in the annotation pool.
Based on the $\roberta_{\text{analysis}}$ model confidence in the \textit{gold label}, we sample 125 examples from the easy subset, as determined by \abr{AFLite}, in a round-robin fashion for each decile between 0 and 1. We repeat this to collect 125 examples from the difficult subset.

\paragraph{Abductive NLI.}
Since the publicly-released $\alpha$-NLI dataset only contains examples that survived adversarial filtering (``difficult'' examples), we reached out to the authors to obtain \textit{easy} examples that were filtered out.
We source our original examples from the test split of $\alpha$-NLI. 
We train a \roberta-large model on examples from the train split of \anli{} and follow the same stratified sampling protocol on 125 examples from each of the easy and difficult subsets, according to the confidence of the $\text{RoBERTa}_{\text{analysis}}$ in the correct label.

\subsection{Paraphrased Example Collection}
\label{sec:mturk}
We obtain 250 examples per dataset (\anli{}, \snli{}, \atomic{}, \social{}), resulting in 1000 examples for which we collect paraphrases. 
\paragraph{Crowdsourcing.} We use Amazon Mechanical Turk to collect paraphrases of the target portions of each example. 
Workers are shown context sentences and must write a paraphrase of the target sentence(s) according to the definition of label-preserving paraphrasing presented earlier.

We abstract the underlying reasoning task away, presenting \anli{} examples as short stories that require paraphrasing middle sentences, and $\delta$-NLI examples as scenarios with weakening or strengthening evidence.
In order to encourage diversity of paraphrases, we display the Jaccard similarity between tokens in the original sentence and the paraphrase as workers typed.
Figure~\ref{fig:interface} shows our annotation interface and instructions for collecting paraphrases of \anli{} and \dnli{} problems respectively.

Workers provide 3 paraphrases of both plausible and implausible hypotheses for \anli{} examples and 3 paraphrases of updates for \dnli{} examples. 
In the case of \anli{}, we randomly pair together plausible and implausible hypotheses written by the same worker to construct paraphrased examples.
Each example was annotated by 3 workers. See Appendix~\ref{appendix:crowdworking} for more details.

\begin{table}[t]
    \centering
    \resizebox{\columnwidth}{!}{%
    \begin{tabular}{ll}
        \toprule
        \rotatebox[origin=c]{90}{\textbf{Example}} &
        \makecell*[{{p{10.7cm}}}]{
            \colorbox{lightpurple}{\textcolor{darkpurple}{\textbf{O1:}}} George decided to buy a TV. \\ 
            \colorbox{lightteal}{\textcolor{teal}{\textbf{$h^+$:}}} He took it home, unpacked it, and placed it where he wanted it. \\
            \colorbox{lightorange}{\textcolor{burntorange}{\textbf{O2:}}} Things had turned out just as he'd hoped.
        }\\
        
        \midrule
        \rotatebox[origin=c]{90}{\textbf{\textcolor{coolgreen}{Accepted}}}
        &\makecell*[{{p{10cm}}}]{
        \checkmarksmall{} George unboxed the TV and placed it on his mantelpiece.\\
        \checkmarksmall{} He removed the plastic and positioned the TV where he had planned to put it.\\
        \checkmarksmall{} George brought the new TV home and mounted it on the wall.}\\

        \midrule
        \rotatebox[origin=c]{90}{\textbf{\textcolor{coolred}{Rejected}}}
        &\makecell*[{{p{10cm}}}]{
        \crossmarksmall{} George came home and drank a cup of coffee.\\
        \crossmarksmall{} He returned the TV.}\\
        
        \bottomrule
    \end{tabular}}
    \caption{An \anli{} example and paraphrases that were accepted and rejected from crowdworkers.}
    \label{tab:crowdworker-examples}
\end{table}

\paragraph{Paraphrase Example Validation.}
Ensuring semantic equivalence between paraphrased examples and original reasoning problems is essential to our study. 
Inadvertent removal of the crux of the reasoning problem while paraphrasing may result in invalid examples.
Table~\ref{tab:crowdworker-examples} includes an \anli{} example with paraphrases that were both accepted and rejected from crowdworkers based on our definition of label-preserving paraphrasing.
The first and third \textcolor{coolgreen}{\textit{accepted}} paraphrases both introduce new pieces of information (``mantelpiece'', ``mounted it on the wall'') but do not violate the situation evoked by the original problem.
In contrast, the first \textcolor{coolred}{\textit{rejected}} paraphrase is incompatible with the situation and the second rejected paraphrase does not retain the plausibility of the hypothesis.

We opt to have an author validate all paraphrases, each within the context of the problem.\footnote{We explored using NLI models to automatically ensure semantic equivalence, but found it too strict of a formulation to capture the spirit of label-preserving paraphrasing.}
A second author and two external annotators annotated a sample of 100 paraphrases, again labeling each as either valid or invalid.
We obtain a Fleiss's Kappa~\cite{fleiss1971measuring} value of $\kappa=0.81$ between all validators---the two authors and two external validators.
This measures agreement \textit{on the criterion of label-preservation}, reflecting whether a paraphrase written by a crowdworker was of high enough quality to admit into our dataset, as opposed to a measurement of agreement on the correct label given paraphrased examples.

\paragraph{Dataset Overview.} Our resulting dataset, \abr{ParaNlu}, contains 1000 examples uniformly split across \anli{}, \snli{}, \atomic{}, and \social{}. Table \ref{tab:paraphrase-dataset-sizes} shows the total number of post-validation paraphrased examples per data split, and the statistics of sizes of buckets.

\section{Consistency on Human Paraphrases}
\NewDocumentCommand{\Frame}{}{\Block[tikz={draw, offset=1pt,line width= 1pt}]{1-2}{}}

\begin{table*}[h!]
\centering
\tiny

\setlength\tabcolsep{3pt}
\begin{NiceTabular}{lcllccc|cllccc|cllccc|cllccc} 
\toprule & \multicolumn{6}{c}{\textbf{\anli{}}}  & \multicolumn{6}{c}{\textbf{\snli{}}}  & \multicolumn{6}{c}{\textbf{\atomic{}}} & \multicolumn{6}{c}{\textbf{\social{}}}  \\
& \accoriginal{}   & \acctest{} & \accparaphrases  & \multicolumn{1}{l}{\pstay} & \cellcolor{blue!7}\accparaphrasesc & \multicolumn{1}{l}{\cellcolor{blue!7}\pstayc} & \multicolumn{1}{l}{\accoriginal} & \acctest & \accparaphrases & \multicolumn{1}{l}{\pstay} & \cellcolor{blue!7}\accparaphrasesc & \multicolumn{1}{l}{\cellcolor{blue!7}\pstayc} & \multicolumn{1}{l}{\accoriginal} & \acctest & \accparaphrases & \multicolumn{1}{l}{\pstay} & \cellcolor{blue!7}\accparaphrasesc & \multicolumn{1}{l}{\cellcolor{blue!7}\pstayc} & \multicolumn{1}{l}{\accoriginal} & \acctest & \accparaphrases & \multicolumn{1}{l}{\pstay} & \cellcolor{blue!7}\accparaphrasesc & \multicolumn{1}{l}{\cellcolor{blue!7}\pstayc}  \\ 
\midrule
\textbf{Lexical (BoW)}   & 44.8 & 52.4 & 44.2 & 100 & \cellcolor{red!25} 52.4 & \cellcolor{red!25}100 & 58.0  & 55.7 & 53.7  & 82.2 & \cellcolor{blue!7} 52.6 & \cellcolor{blue!7} 80.9 & 49.2 & 53.6  & 51.4 & 76.5 & \cellcolor{blue!7}53.2 & \cellcolor{blue!7}76.5  & 57.6 & 61.8  & 51.4  & 78.2 & \cellcolor{blue!7}54.9 & \cellcolor{blue!7}78.5  \\
\textbf{BiLSTM}  & 53.5 & 51.6 & 54.2 & 100 & \cellcolor{blue!7} 51.6 & \cellcolor{blue!7}100 & 62.0 & 68.0  & 57.6  & 73.2 & \cellcolor{blue!7} 60.4 & \cellcolor{blue!7} 74.2 & 52.8 & 67.4 & 54.2 & 73.1 & \cellcolor{blue!7}61.1 & \cellcolor{blue!7}74.8  & 60.4 & 72.0 & 52.4  & 71.7 & \cellcolor{blue!7}59.7 & \cellcolor{blue!7}72.8  \\
\textbf{RoBERTa} & 53.6 & 83.5 & 56.4 & 69.8 & \cellcolor{red!25}81.5 & \cellcolor{red!25}86.3   & 51.2 & 86.7  & 53.8  & 74.8 &  \cellcolor{red!25}84.6  &  \cellcolor{red!25}90.1 & 53.6  & 82.6 & 54.8 & 76.2 & \cellcolor{blue!7}77.9 & \cellcolor{blue!7}87.1 & 51.6 & 90.9   & 56.9  & 74.3 & \cellcolor{red!25}87.8 & \cellcolor{red!25}91.9 \\
\textbf{DeBERTa-V3} & 85.6 & 90.6 & 73.5  & 78.4 & \cellcolor{blue!7}77.3 & \cellcolor{blue!7}79.7  & 76.8 & 91.2  & 70.4 & 82.8 & \cellcolor{blue!7}80.5 & \cellcolor{blue!7}84.1  & 70.4  & 88.1  & 66.8 & 82.7 & \cellcolor{red!25}81.8 & \cellcolor{red!25}87.3  & 78.4 & 94.1 & 71.9 & 82.2 & \cellcolor{blue!7}78.6 & \cellcolor{blue!7}83.7  \\
\textbf{Unified RoBERTa} & —    & —  & —  & —   &\cellcolor{blue!7} —  &\cellcolor{blue!7} —  & 66.0   & 85.7  & 59.9 & 78.0 & \cellcolor{blue!7}81.5 & \cellcolor{blue!7}88.6  & 65.6 & 84.5 & 62.8  & 80.7 & \cellcolor{blue!7}78.8 & \cellcolor{blue!7}86.8  & 70.8 & 90.4  & 65.7  & 77.7 & \cellcolor{blue!7}83.3 & \cellcolor{blue!7}87.7 \\
\textbf{GPT-3 Curie} & 46.8 & 53.5 & 46.2  & 80.1 & \cellcolor{blue!7}51.4 & \cellcolor{blue!7}79.1 & 52.8 & 48.9  & 51.1 & 89.3 & \cellcolor{blue!7}51.3 & \cellcolor{blue!7}88 & 52.0 & 49.5 & 52.8 & 90.2 & \cellcolor{red!25}50.6 & \cellcolor{red!25}89.6 & 55.6  & 49.9 & 57.4  & 91.3 & \cellcolor{blue!7}53.0 & \cellcolor{blue!7}90.9 \\
\bottomrule
\end{NiceTabular}
\caption{Consistency across modeling architectures. \accoriginal{} is accuracy on original examples in \paranlu{}, \acctest{} is full test set accuracy, \accparaphrases{} is accuracy on paraphrased examples (\accparaphrasesc{} is corrected), and \pstay{} is paraphrastic consistency (\pstayc{} is corrected). No model achieves both high \accparaphrasesc{} and \pstayc{} (columns highlighted in blue). Optimizing \accparaphrasesc{} may come at the cost of \pstayc{}, or vice versa. We highlight in red the set of \textit{Pareto-optimal} points, or those that are not strictly dominated in \accparaphrasesc{} or \pstayc{} by any other model.}
\label{tab:modeling-paradigms-consistency}
\end{table*}

We first examine several different models' behavior on \paranlu{} to measure robustness to different linguistic expressions.
While language models such as \roberta{} are trained on vast amounts of text that may instill some paraphrastic consistency, especially given label-preserving paraphrases that not be semantically equivalent, other non-pretrained models without access to such knowledge may falter.
We characterize the progress of models with respect to \pstay{} to understand whether factors such as training setups (training from scratch, supervised finetuning, prompting) and model complexity (ranging from bag-of-words representations to GPT-3) affect consistency.

\subsection{Model Variants}
\label{sec:model-variants}
We train 5 different types of models per data source.
For all models, we use the same set of examples that were used to finetune the $\roberta{}_{\text{analysis}}$ models introduced in \S\ref{sec:example-selection}.

\paragraph{Bag of Words.} We train bag-of-words models (BoW) using \texttt{fasttext}, an off-the-shelf text classification library, with a maximum of 4-grams~\cite{joulin-etal-2017-bag} for 5 epochs with the default learning rate of 0.1.

\paragraph{\bilstm{}.} We train end-to-end BiLSTM models using the architecture from~\citet{conneau-etal-2017-supervised} and initialize them with \abr{GloVe} embeddings~\cite{pennington-etal-2014-glove}. 
We use 3 fully connected layers for classification with max pooling. 
After tuning on the development sets, models are trained for 10 epochs with early stopping and a batch size of 64.

\paragraph{\roberta{}.} We use the $\text{RoBERTa}_\text{analysis}$ models in \S\ref{sec:example-selection}, and add one more setting for defeasible examples in which we finetune a \roberta{}-large model on all combined data across the 3 data sources in \dnli{}, which we refer to as a unified \roberta{} model. 
All \roberta{}-large models were finetuned for 2 epochs with a learning rate of 2e-5 and a batch size of 32.

\paragraph{\deberta{}.} We finetune \deberta-v3-large~\cite{he2021debertav3} for 2 epochs with a learning rate of 5e-6 and a batch size of 16.

\begin{table}[t]
    \centering
    \resizebox{\columnwidth}{!}{%
    \begin{tabular}{ll}
        \toprule
        \rotatebox[origin=c]{90}{\textbf{\anli{}}} &
        \makecell*[{{p{10.7cm}}}]{
            \texttt{Given a story with a \textcolor{darkpurple}{\textbf{beginning}} and an \textcolor{burntorange}{\textbf{end}}, select from two choices the \textcolor{teal}{\textbf{middle}} sentence that is most plausible and best explains the \textcolor{darkpurple}{\textbf{beginning}} and \textcolor{burntorange}{\textbf{end}} of the story.}
        }\\
        \midrule
        \rotatebox[origin=c]{90}{\textbf{\dnli{}}}
        &\makecell*[{{p{10.7cm}}}]{
        \texttt{Given a \premise{} sentence, a \hypothesis{} sentence is defeasible if there exists an \update{} sentence (consistent with the \premise{}) such that a human would find the \hypothesis{} less likely to be true after learning the \update{}. An \update{} is called a weakener (abbreviated W) if given a \premise{} and \hypothesis{}, a human would most likely find the \hypothesis{} less likely to be true after learning the \update{}; if they would find the \hypothesis{} more likely to be true, then the \update{} is called a strengthener (abbreviated S). Given a Premise, a \hypothesis{}, and an \update{} sentence, assign either W or S to the \update{} sentence.}}\\
        
        \bottomrule
    \end{tabular}}
    \caption{Few-shot prompts for \anli{} and \dnli{} tasks. Both prompts were presented to \gpt{} along with 36 in-context examples.}
    \label{tab:gpt3-prompts}
\end{table}

\paragraph{\gpt{}.} Lastly, we experiment with prompting GPT-3~\cite{brown2020language} using \abr{text-curie-001} (prompts in Table~\ref{tab:gpt3-prompts}).
For \anli{}, we randomly sample 36 examples from the training set and include instructions derived from those shown to the crowdworkers that annotated the \anli{} dataset. For \dnli{}, we randomly sample 12 examples per dataset (36 in-context examples total) and include the task definition from~\citet{rudinger-etal-2020-thinking} in the prompt.
Since we cannot reliably extract a softmax distribution over binary classes for our tasks (\gpt{} is not a classification model), we calculate model confidence in a particular class by extracting log probabilities associated with the tokens for both labels and normalize them. 

\subsection{Results}

\begin{figure}[t]
\centering
\includegraphics[scale=0.53]{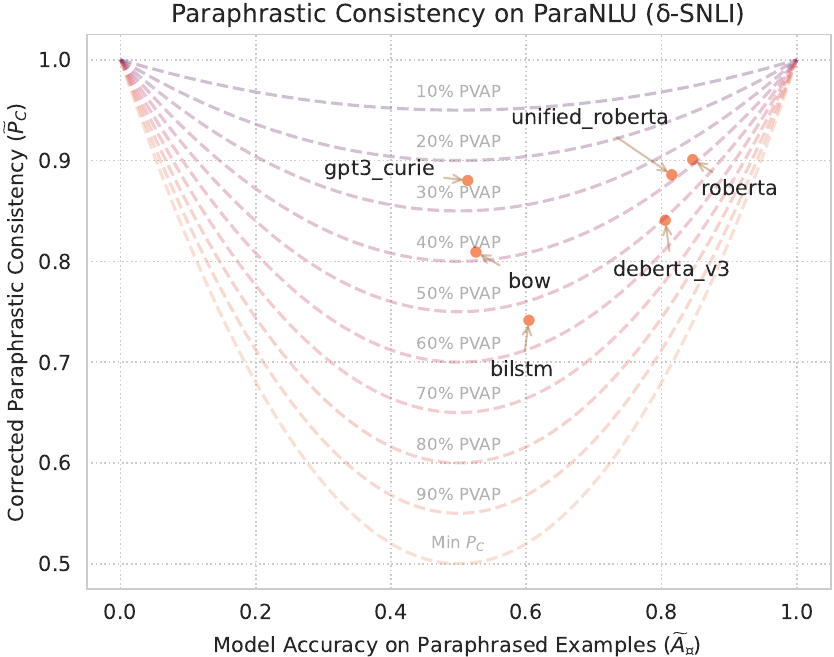}
\caption{Paraphrastic consistency (\pstayc{}) of different models on \snli{} paraphrased examples. All models still have room for improvement in \pstayc{} for their accuracy level. Here, we add supporting lines to denote varying levels of the proportion of variance attributable to paraphrasing, or \pvap{}.}
\label{fig:modeling-paradigms-pstay}
\end{figure}

For all models, we compute \pstay{} (Equation~\ref{eq:pstay}). In addition, we undo the biasing effects of the stratified sampling according to model confidence (\S\ref{sec:example-selection}) and report a corrected version of paraphrastic consistency, \pstayc{}, by weighting the expectations in Equation~\ref{eq:pstay} according to the distribution of model confidences in the correct label of the corresponding test set. We compute four accuracy metrics: (1) accuracy on original examples in the \paranlu{} (\accoriginal{}), (2) accuracy on the test set of the original dataset (\acctest{}), (3) accuracy on all paraphrases across all buckets (\accparaphrases{}), and (4) \textit{corrected} accuracy on all paraphrases across all buckets (\accparaphrasesc{}) which is weighted in the same manner as \pstayc{} to undo stratification effects.

Table~\ref{tab:modeling-paradigms-consistency} shows these accuracy metrics along with \pstay{} and \pstayc{} for all models across all data sources.
We highlight \accparaphrasesc{} and \pstayc{}, as they are \textit{complementary} metrics meant to jointly assess model performance. 
Some models optimize \accparaphrasesc{} at the cost of \pstayc{}, or vice versa. 
To capture models that balance both, we highlight (\accparaphrasesc{}, \pstayc{}) points that are \textit{Pareto-optimal} for each dataset.
The highest performing model according to \accparaphrasesc{}, \roberta{}, earns a \pstayc{} of around 0.9, indicating room to improve on its paraphrastic consistency \textbf{for its accuracy level}.
We observe that a \gpt{}-\abr{curie} model with minimal prompt engineering (we simply use the definition of defeasible inference directly from~\citet{rudinger-etal-2020-thinking}) along with a handful of in-context examples has a \pstayc{} value competitive with a \roberta{} model finetuned on thousands of examples.
A stronger \gpt{} variant may better perform defeasible reasoning while maintaining a competitive \pstay{}.

Figure~\ref{fig:modeling-paradigms-pstay} visualizes the relationship between accuracy and \pstayc{} for models on \snli{} examples (Figure~\ref{fig:remaining-modeling-pstay} shows similar plots for \atomic{} and \social{} examples). 
For each model, we plot \accparaphrasesc{} on the x-axis and \pstayc{} on the y-axis along with two types of supporting curves for the \snli{} split of \paranlu{}. 
The curve with the lowest minima (labeled \texttt{Min \pstay{}}) indicates the theoretical lower bound for \pstay{} \textbf{given a particular accuracy level}: if all the variance in model correctness (Equation~\ref{eq:total-variance-law}) is attributable to the paraphrasing variance term present in Equation~\ref{eq:uv-pstay}, then the minimum possible value for \pstay{} is $1 - 2 * (Acc * 1 - Acc)$, where $Acc * (1 - Acc)$ is the variance of a Bernoulli random variable with probability $Acc$.
In addition to this theoretical lower bound, we plot curves indicating the proportion of total variance attributed to paraphrasing (labeled \texttt{\%PVAP}). As \pstay{} increases, less variance is attributed to variance within buckets due to phrasing.
Visualizing this relationship makes it clear that accuracy alone provides an incomplete picture of model performance.
A perfect model would reside in the top right of Figure~\ref{fig:modeling-paradigms-pstay}, not only achieving high accuracy but also high paraphrastic consistency.
Models with similar accuracies may have largely different \pstayc{} values, indicating to practitioners how sensitive they are to problem phrasing. 

We now turn to a series of experiments to better characterize paraphrases in \paranlu{} as well as contributing factors to model's \pstay{} value using our best performing model, \roberta{}.

\label{sec:model-consistency}

\section{Paraphrase Source}
\label{sec:auto-vs-human}
\begin{figure}[t]
\centering
\includegraphics[scale=0.53]{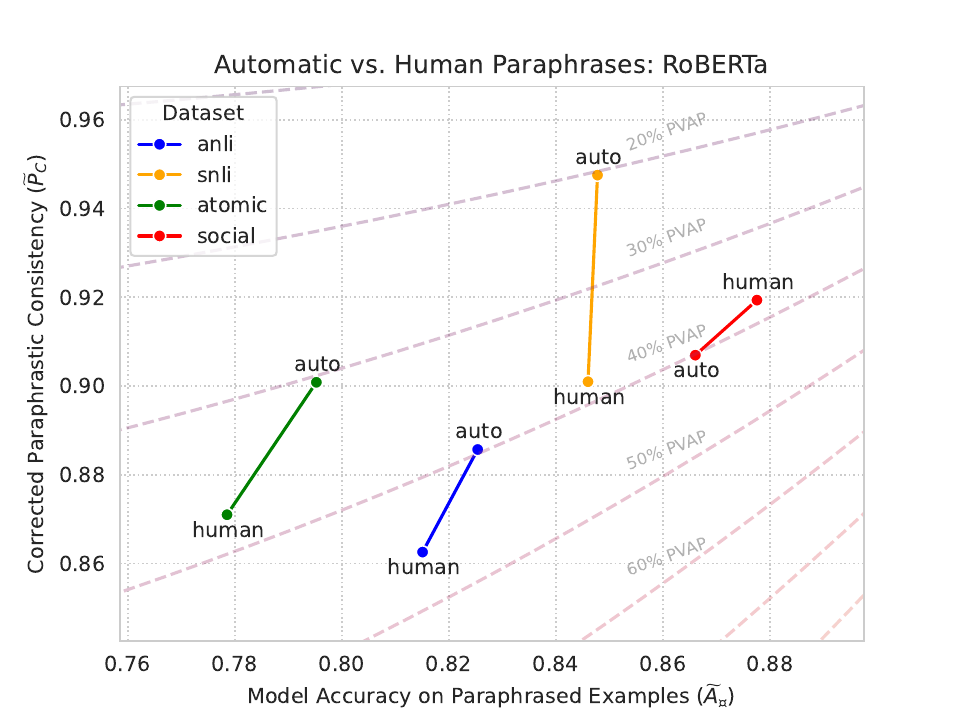}
\caption{\roberta-large model on automatic versus human paraphrases. Models are more consistent on automatic than human paraphrases. Dashed lines indicate varying levels of \pvap{}.}
\label{fig:auto-v-human-plot}
\end{figure}

Human-written paraphrases in \abr{ParaNlu} span all of the transformations delineated by~\citet{bhagat2013paraphrase}, sometimes involving more complex reasoning that falls between linguistic and world knowledge.
Such paraphrases were elicited by providing humans the entire example, encouraging them to engage with both the reasoning problem itself and the wide scope of possible meaning-preserving transformations.
To understand the utility of label-preserving paraphrases, we compare a \roberta{} model's behavior on our human-written paraphrases with paraphrases generated automatically, as previous studies have explored~\cite{verma-etal-2023-evaluating}.

Using our \roberta{} models (\S\ref{sec:model-variants}), we probe the relationship between paraphrastic consistency (\pstayc{}) and the \textit{source} of paraphrased examples. 
Are models more robust to the paraphrastic transformations produced by automatic paraphrase generation models, or do they only struggle with the more complex, example-aware transformations made by humans?

Since human paraphrases and model paraphrases are generated by different processes, and are thus drawn from different distributions, they may exhibit different properties.
Paraphrase generation models are predisposed to biases arising from n-gram frequency effects.\footnote{For example, a generation model is less likely to paraphrase \textit{``The camera zooms out to show the man spraying the car with soap''} to \textit{``The camera zooms out to servicemen sprinkling the automobile with soap''}.}
However, reasoning models should exhibit consistency \textit{regardless} of whether correct answers are phrased as high-probability sentences under a language model.

We use two models (\S\ref{sec:paraphrase-models}) to automatically paraphrase target sentences in original examples and compare model \pstay{} on automatic and human paraphrases.

\subsection{Experimental Setup}
For each original example in \abr{ParaNlu}, we sample paraphrases of targets from generation models. 
As with humans, we elicit paraphrases of target sentences: update sentences for \dnli{} and both hypothesis sentences for \anli{}. 
In contrast to the human elicitation process, however, we do \textit{not} provide any context sentences to generation models.
While this limits the scope of possible paraphrases, it allows us to gauge the value of exposure to context during paraphrasing.

We adopt a generate-then-validate scheme and have an author again validate all target paraphrases to ensure that their consistency with our definition of label-preserving paraphrasing.
In the case of \anli{} examples, where there are multiple target sentences, we randomly pair valid paraphrases together, resampling where necessary when numbers of valid generated hypotheses are unequal.

\subsection{Paraphrase Generation Models}
\label{sec:paraphrase-models}

\paragraph{Quality-Controlled Paraphrase Generation.} We use a \qcpg{} model~\cite{bandel-etal-2022-quality}, a controllable paraphrase generation system that conditions on a 3-dimensional vector encoding semantic similarity and lexical and syntactic distances. We pool all paraphrases from a per-example sweep of these hyperparameters.

\paragraph{\gpt{}.} In addition to a supervised, explicitly controllable paraphrase generation model, we elicit paraphrases from \gpt{} using 10 in-context examples of paraphrases randomly sampled from \paranlu{}. 
Setting temperature to 0.7, we sample 9 paraphrases from \abr{text-davinci-002} per target sentence from original examples and similarly validate them to ensure label preservation. 

\begin{table}
\centering
\tiny
\resizebox{\columnwidth}{!}{%
\begin{tabular}{lllll} 
\toprule
& & \multicolumn{1}{c}{\textbf{lex}} & \multicolumn{1}{c}{\textbf{syn}} & \multicolumn{1}{c}{\textbf{sem}}\\
\cline{3-5}
\multirow{2}{*}{\textbf{\anli{}}}  & \textit{automatic}  & 25.0 & 20.3 & 74.3  \\
& \textit{human} &  \textbf{35.3} &  \textbf{26.8} &  \textbf{64.0} \\ 
\hline
\multirow{2}{*}{\textbf{\snli{}}} & \textit{automatic}  & 24.1 & 18.5 & 76.3 \\
& \textit{human} &  \textbf{34.4} &  \textbf{24.6} &  \textbf{67.4} \\ 
\hline
\multirow{2}{*}{\textbf{\atomic{}}} & \textit{automatic}  & 30.4 & 19.2 & 66.7 \\
& \textit{human} &  \textbf{36.9} &  \textbf{22.4} &  \textbf{58.7} \\ 
\hline
\multirow{2}{*}{\textbf{\social{}}} & \textit{automatic}  & 30.8 & 22.2 & 70.1 \\
& \textit{human} & \textbf{40.0} & \textbf{24.5} & \textbf{60.1} \\
\bottomrule
\end{tabular}}
\caption{Lexical (\textbf{lex}) and syntactic (\textbf{syn}) diversity of human and automatic paraphrases and semantic similarity (\textbf{sem}) as compared to original target sentences. Human paraphrases are more diverse than automatic ones: they exhibit higher lexical and syntactic diversity and lower semantic similarity on all datasets (bolded).}
\label{tab:surface-form}
\end{table}

\subsection{Results}
In total, we generate 7,295 valid paraphrased examples across 1000 examples by pooling together all \textit{valid} examples from both \qcpg{} and \gpt{} and evaluate our \roberta{} models on these examples.
Figure~\ref{fig:auto-v-human-plot} plots \accparaphrasesc{} on the x-axis and \pstayc{} on the y-axis for human-generated and automatically-generated paraphrased examples for each dataset.
On all datasets with the exception of \social{}, we observe that models have a higher \pstayc{} value on automatically generated paraphrased examples than on human-elicited paraphrases.
We hypothesize that this pattern may not hold for \social{} due to the fact that the dataset does not contain premise sentences, and hence has a smaller scope for more complex transformations involving context.
This result suggests that reasoning models may be more robust to the simpler, in-distribution types of paraphrase transformations that automatic paraphrase generation models produce than to those written by human annotators, indicating that over-reliance on evaluation using synthetically generated data may be misleading.

\begin{table}
\centering
\resizebox{\columnwidth}{!}{
\begin{tabular}{llccc|ccc|cc} 
\toprule
&   & \multicolumn{3}{c|}{\textbf{partial-input}}  & \multicolumn{3}{c|}{\textbf{full-input}}  & \multicolumn{1}{l}{} & \multicolumn{1}{l}{}                        \\
& artifacts...  & \multicolumn{1}{l}{\accoriginal{}} & \multicolumn{1}{l}{\accparaphrases{}} & \multicolumn{1}{l|}{\accparaphrasesc{}} & \multicolumn{1}{l}{\accoriginal{}} & \multicolumn{1}{l}{\accparaphrases{}} & \multicolumn{1}{l|}{\accparaphrasesc{}} & \multicolumn{1}{l}{\pstay{}} & \multicolumn{1}{l}{\pstayc{}}  \\ 
\cline{2-10}
\multirow{2}{*}{\textbf{\snli{}}}   & \textit{likely}    & 100 & 81.8 & 54.6 & 54.3 & 56.6  & 85.8 & 74.7 & 91.4 \\
& \textit{unlikely} & 0 & 20.7 & 6.9 & 45.5 & 48.8  & 79.6 & 75.1 & 84.0 \\ 
\hline
\multirow{2}{*}{\textbf{\atomic{}}} & \textit{likely} & 100  & 77.6  & 53.9 & 55.6 & 58.3 & 78.0 & 76.4 & 87.4 \\
& \textit{unlikely} & 0  & 21.3   & 8.3 & 50.9 & 49.9 & 79.2 & 75.9 & 86.5  \\ 
\hline
\multirow{2}{*}{\textbf{\social{}}} & \textit{likely} & 100  & 77.2 & 58.9  & 52.9 & 55.5 & 85.8 & 73.9 & 90.9 \\
& \textit{unlikely} & 0 & 28.8  & 8.4 & 50 & 58.7 & 90.7 & 74.7  & 93.5\\
\bottomrule
\end{tabular}}
\caption{Paraphrastic consistency on examples which are likely and unlikely to contain artifacts, as predicted by a partial-input baseline. Inconsistency on both types of examples indicates \pstay{} is attributable to factors \textit{beyond} artifacts.}
\label{tab:artifacts}
\end{table}

\paragraph{Paraphrase Diversity.} To dissect this result further, we measure lexical diversity, syntactic diversity, and semantic similarity~\cite{bandel-etal-2022-quality} of target paraphrases and original target sentences.
\textit{Lexical distance} is measured by the normalized character-level minimal edit distance between the bag of words~\cite{bandel-etal-2022-quality}, and \textit{syntactic distance} is computed as the normalized tree edit distance between the third level constituency parse trees of the original target and the paraphrased target~\cite{iyyer2018adversarial}.
We measure semantic similarity using BLEURT~\cite{sellam-etal-2020-bleurt} as in~\citet{bandel-etal-2022-quality}.
Across all four data splits in \paranlu{}, human-elicited paraphrases are more lexically and syntactically diverse, as well as less semantically similar to original examples than automatically generated paraphrases (Table~\ref{tab:surface-form}).
In addition, we find that automatically generated paraphrases are 3--4\% more likely to be bidirectionally entailed than human-written paraphrases, as detected by a \roberta-large model finetuned on SNLI~\cite{bowman-etal-2015-large}, MNLI~\cite{williams-etal-2018-broad}, ANLI~\cite{nie-etal-2020-adversarial}, and FEVER~\cite{thorne-etal-2018-fever}.

Taken together, these results underscore the benefit of our human annotation to generate \paranlu{} --- evaluation solely on automatically generated paraphrases, as others have done, is insufficient to fully characterize their robustness.

\section{Do artifacts explain inconsistency?}
Many \abr{NLP} datasets are constructed by crowdworkers writing most or all parts of a natural language reasoning problem.
While efficient and scalable, this paradigm can give rise to annotation artifacts~\cite{gururangan-etal-2018-annotation}, or statistical biases in parts of examples that correlate with the correct label~\cite{mccoy-etal-2019-right}.
For example, \citet{gururangan-etal-2018-annotation} found that negation words (\textit{``no''}, \textit{``nothing''}, etc) in NLI examples are strong indicators of the contradiction label.

One way to detect artifacts in datasets is through a partial-input baseline~\cite{poliak-etal-2018-hypothesis, feng-etal-2019-misleading}, a setting in which only the \textit{target} portion of an NLI instance (i.e the hypothesis in a traditional NLI setup) is used to train a model to predict entailment.
When partial-input models achieve high accuracy, it is often indicative of annotation artifacts.

Full-input models that are trained on datasets with annotation artifacts may learn to rely on such shallow signals instead of performing true inferential reasoning. 
This may lead to lower paraphrastic consistency in the face of alternative phrasings of examples, since they may no longer contain the annotation artifact the model leveraged.
Can we attribute a model's issues with paraphrastic consistency on \paranlu{} entirely to the presence of annotation artifacts?
That is, are models only inconsistent on examples with artifacts, since they are relying on spurious correlations? Or, are they inconsistent even when no artifacts are present?

\subsection{Experimental Setup}
\citet{rudinger-etal-2020-thinking} find that partial-input baselines trained on \dnli{} perform at least 10\% better than random chance, indicating the presence of annotation artifacts in their dataset.
As such, we focus on \dnli{} for our experiments.\footnote{We choose \dnli{} instead of \anli{} for this experiment, since the authors of \anli{} released their dataset \textit{after} running adversarial filtering.} 
For each split of \dnli{}, we train a \roberta-large model using only the update sentence as input, keeping the training hyperparameters identical to the full-input \roberta{} models from \S\ref{sec:model-variants}.
Then, we use these partial-input models to partition buckets in \paranlu{} into two subsets: those on which the partial-input model correctly predicted the label from the update sentence of the \textit{original example}, indicating that an artifact is likely present in the original example, and those incorrectly predicted.
Using the full-input \roberta{} models from \S\ref{sec:model-variants}, we then compute \pstayc{} on both example subsets, and the accuracy of partial-input and full-input models on paraphrased and original examples.

\subsection{Results}

Table~\ref{tab:artifacts} shows the accuracy metrics for both partial-input and full-input models on original and paraphrased examples in \paranlu{}, as well as paraphrastic consistency metrics on both examples that are likely ($+$) and unlikely ($-$) to contain artifacts.
While not all original examples that a partial-input model predicts correctly \textit{necessarily} have artifacts, we expect that (1) examples with particularly strong artifacts are grouped in the \textit{likely} ($+$) category, and (2) the \textit{unlikely} ($-$) category contains a significantly smaller number of examples with strong artifacts.
We observe a dramatic drop in the accuracy of a partial-input baseline on original examples (\accoriginal{}) and paraphrased examples (\accparaphrasesc{}), indicating that most artifacts detectable with a partial-input model \textit{do not} project through our label-preserving paraphrase process.

Even on examples \textit{unlikely} ($-$) to contain artifacts, where a full-input model cannot rely on shallow signals, models do still have room to improve their paraphrastic consistency.
These results indicate that issues with paraphrastic consistency are attributable to factors beyond the presence of artifacts in examples.

\section{Training Dynamics and Paraphrastic Consistency}
Lastly, we explore the relationship between different parts of the model training pipeline (e.g. pretraining and finetuning) and paraphrastic consistency.
How does consistency \textit{change} as these training processes progress, and does it change in a similar manner as accuracy? 
Is it the case that simply increasing the volume of pretraining or finetuning data linearly impacts paraphrastic consistency?
We train a series of \roberta{} models and adjust the number of pretraining tokens (\S\ref{sec:pretraining}) and finetuning examples (\S\ref{sec:finetuning}) to explore how they impact a model's consistency.

\subsection{Pretraining and \pstay}
\label{sec:pretraining}

\paragraph{Experimental Setup.} Using the \abr{MiniBERTa}s~\cite{warstadt-etal-2020-learning}, a series of \roberta{} models pretrained from scratch on varying numbers of tokens, we compare models trained on 1M, 10M, 100M, and 1B tokens along with \roberta{}-base, which is pretrained on approximately 30B tokens.
All models have the same number of parameters as \roberta{}-base (125M), with the exception of the model trained on 1M pretraining tokens which was scaled down in accordance with the smaller volume of pretraining data, and has 45M parameters.
We \textit{finetune} all models on the same data (\S\ref{sec:example-selection}) and keep all hyperparameters constant (batch size of 64, 2 finetuning epochs, and a learning rate of 5e-6), ensuring direct measurement of the impact of pretraining words on paraphrastic consistency without other confounds.

\begin{figure}[t!]
\centering
\includegraphics[scale=0.53]{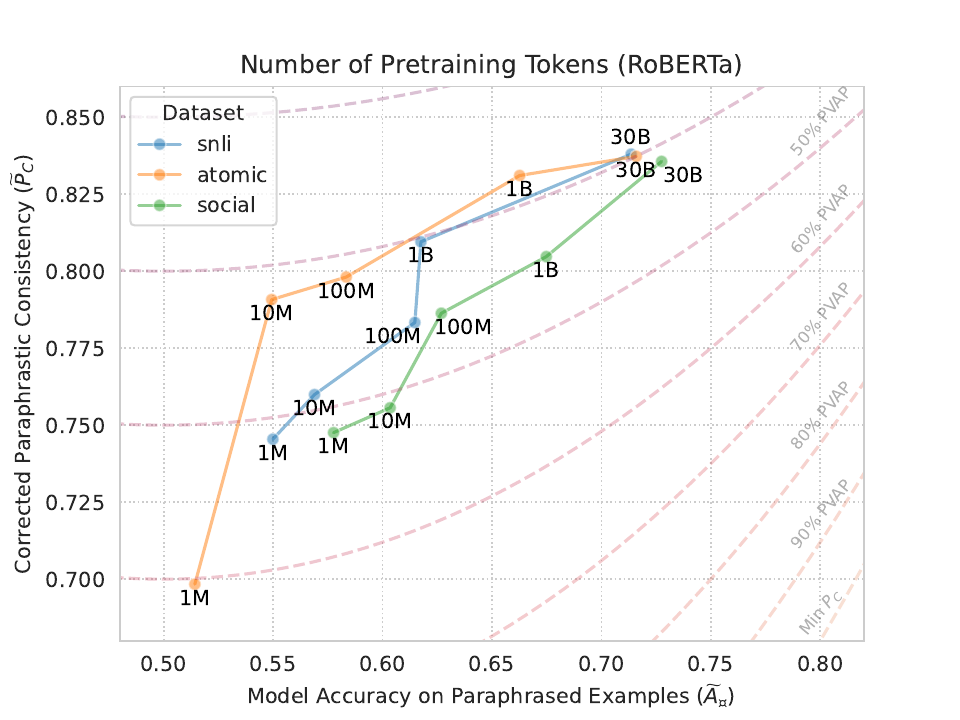}
\caption{Paraphrastic consistency monotonically increases as a model sees more pretraining tokens, but grows rapidly during early pretraining.}
\label{fig:pretraining}
\end{figure}

\paragraph{Results.} Figure~\ref{fig:pretraining} plots model accuracy on paraphrases (\accparaphrasesc{}) against paraphrastic consistency (\pstayc{}), along with the same supporting curves as in Figure~\ref{fig:modeling-paradigms-pstay} corresponding to decreasing proportions of variance attributable to paraphrasing (\abr{pvap}).
As expected, pretraining on increasing amounts of data yields both monotonically increasing accuracy and paraphrastic consistency. 
However, paraphrastic consistency grows more rapidly in the beginning (1M - 100M tokens) as the plots climb steeply between supporting curves and eventually hugs a single \abr{pvap} curve past 100M tokens, indicating a slower payoff of more pretraining tokens.

\subsection{Finetuning and \pstay}
\label{sec:finetuning}

After pretraining, models are endowed with the ability to represent natural language inputs but do not know how to perform a particular reasoning task. 
As such, we expect monotonically increasing accuracy as the model is shown a larger volume of finetuning examples. 
However, it is unclear how \textit{paraphrastic consistency} changes as the model is exposed to more task-specific examples.

\paragraph{Experimental Setup.}
For each dataset, we finetune a series of fully-pretrained \roberta-large models on 1\%, 5\%, 10\%, 50\%, and 100\% of examples from the training split, sampled at random. \atomic{} has 28.3K training examples, \snli{} has 75.2K training examples examples and \social{} has 65.3K examples.
We sample at the premise-hypothesis level and include all examples that share the same premise and hypothesis to prevent data leakage during evaluation.
We hold all training hyperparameters constant, keeping the same configuration as the finetuning in \S\ref{sec:pretraining}.

\begin{figure}[t!]
\centering
\includegraphics[scale=0.53]{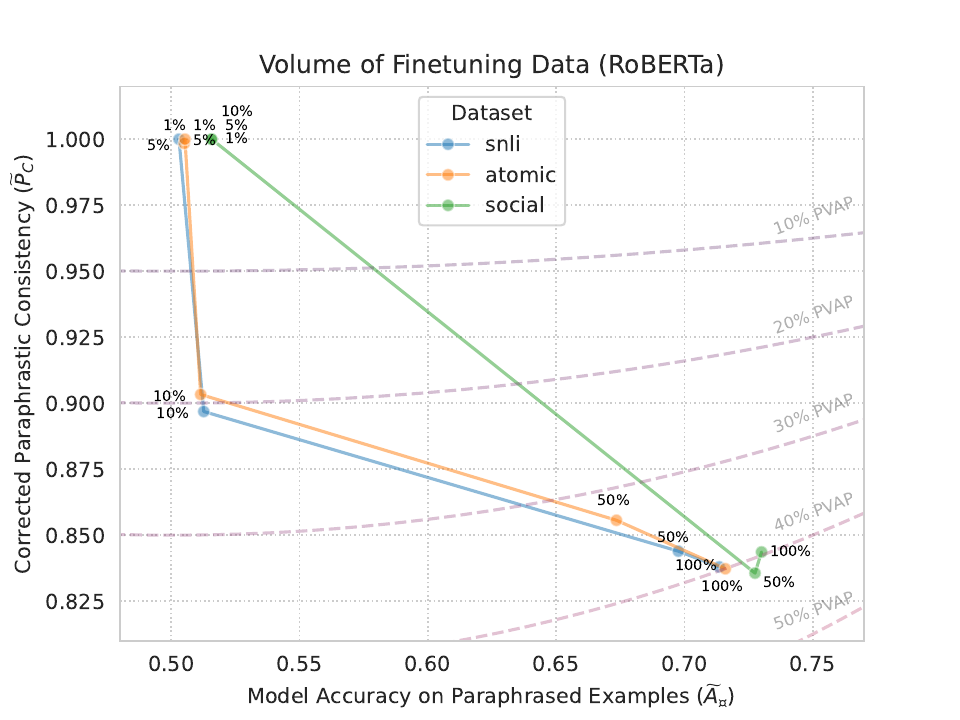}
\caption{Paraphrastic consistency decreases as a model learns increasingly complex decisions boundaries by seeing more finetuning examples.}
\label{fig:finetuning}
\end{figure}

\paragraph{Results.} Figure~\ref{fig:finetuning} plots corrected paraphrase accuracy (\accparaphrasesc{}) against corrected paraphrastic consistency (\pstayc{}) for models trained on increasing numbers of finetuning examples across all three datasets in \dnli{}. 
We observe that as the model starts to learn the task at hand and draw increasingly complex decision boundaries, it is more likely to be inconsistent. 
Models trained on $\leq$5\% of the available training examples are highly consistent since they make the same prediction for all examples (thus earning an accuracy of around 50\%). 
As the the model is shown more examples, it makes finer-grained distinctions between examples, in turn impacting its paraphrastic consistency.
Though our results show this decrease, it is possible that with even more finetuning data, a model's paraphrastic consistency will start to increase again.
This relationship may also be altered if, during finetuning, models are shown increasing amounts of automatically-generated paraphrased examples in order to learn both the task \textit{and} paraphrastic consistency.

\section{Related Work}
Natural language understanding models may produce different predictions in the face of varying expressions of reasoning problems.
A wide range of data generation frameworks have been proposed to study these behaviors in NLP systems.
\citet{iyyer2018adversarial} automatically generate paraphrases with specified syntactic templates and measure accuracy on these adversarial examples. 
\citet{verma-etal-2023-evaluating} introduce a test set of paraphrases generated with a finetuned T5 model~\cite{raffel2020exploring} and measure the accuracy of several models.
\citet{hu-etal-2019-improved} generate paraphrases of MNLI examples using lexical constraints and evaluate an NLI model on the paraphrased inputs, finding that paraphrasing leads to degraded accuracy.
\citet{arakelyan2024semantic} measure the semantic sensitivity of NLI models by automatically generating examples with FLAN-T5~\cite{chung2022scaling} and verifying the generations with bidirectional entailment predicted by pretrained NLI models. 
While scalable, our findings illustrate that it is insufficient to evaluate models on automatic paraphrases alone, as human-written paraphrases introduce more semantic and pragmatic diversity (Section~\ref{sec:auto-vs-human}). 
Moreover, we show that bidirectional entailment as a verification method for generated paraphrases is extremely stringent, precluding us from testing consistency in the face of more challenging label-preserving transformations.

Another body of research studies the creation of adversarial examples to improve model robustness. 
\citet{nie-etal-2020-adversarial} construct an NLI benchmark, Adversarial NLI, by developing a model-in-the-loop framework to iteratively produce examples that models cannot correctly solve.
\citet{naik-etal-2018-stress} develop a suite of adversarial examples to ``stress test'' common failure modes of NLI models, such as phenomena word overlap or negation.
In contrast with these studies, our goal is not to generate a test suite of difficult examples that ``break'' models~\cite{glockner2018breaking}, but rather to carefully measure the role of \textit{paraphrastic} variability in model performance. 

Other approaches to measuring robustness also include counterfactual example generation~\cite{srikanth-rudinger-2022-partial, kaushik2020learning}. 
\citet{kaushik2020learning} recruit humans to create counterfactual examples by minimally editing example text in order to flip the gold label and show that models trained on the original datasets perform poorly on counterfactually-manipulated data.
Similarly, \citet{gardner-etal-2020-evaluating} argue for the creation of evaluative contrast sets, or manual minimal perturbations of dataset examples that change the gold label, in order to probe the decision boundary of models.
Our work has a related, but distinct, counterfactual flavor: if an original annotator had chosen to phrase the question differently \textit{with the same target label}, what is the probability that a model's prediction would stay consistent?
We aim to estimate, in expectation, the reliability of models when they are faced with different phrasings of the same problem.

Most of these studies measure \textit{accuracy} on adversarial examples as the main determination of robustness. 
\citet{elazar-etal-2021-measuring} instead measure the \textit{consistency} of models with respect to factual knowledge, evaluating whether extracted information from masked language models is invariant to paraphrasing using an agreement-based consistency metric. 
Our study is similarly concerned with consistency, however we make precise the relationship between accuracy and consistency on natural language reasoning tasks.

\section{Conclusion}
As more studies investigate the capabilities of LLMs, the ability to disentangle the effects of paraphrastic variability from other target attributes will be an important analytical tool. 

This work introduces a new methodology and dataset for measuring paraphrastic consistency, or \pstay{}, of models on natural language reasoning tasks.
\pstay{} captures the probability that a model will remain consistent in its prediction given different phrasings of the same underlying reasoning problem.
We design \pstay{} as a metric complementary to accuracy, and propose practitioners use it alongside accuracy when diagnosing modeling errors, summarizing a model's performance, or deciding when a model is ready for deployment to users for a particular application.

Our results confirm that paraphrastic sensitivity is present in all models, but decreases with pretraining volume. 
Because \pstay{} only requires model predictions to be labeled as correct or incorrect, our approach can generalize to any task with binary scoring (and where answers must be invariant to paraphrases). 
Future work may consider adapting this approach for tasks with more complex or open-ended evaluations.

\bibliography{tacl2021}
\bibliographystyle{acl_natbib}

\afterpage{\clearpage}
\newpage
\appendix

 \begin{figure}[H]
 \begin{minipage}{\textwidth}
    \begin{subfigure}[b]{\linewidth}
        \centering
        \includegraphics[width=\textwidth]{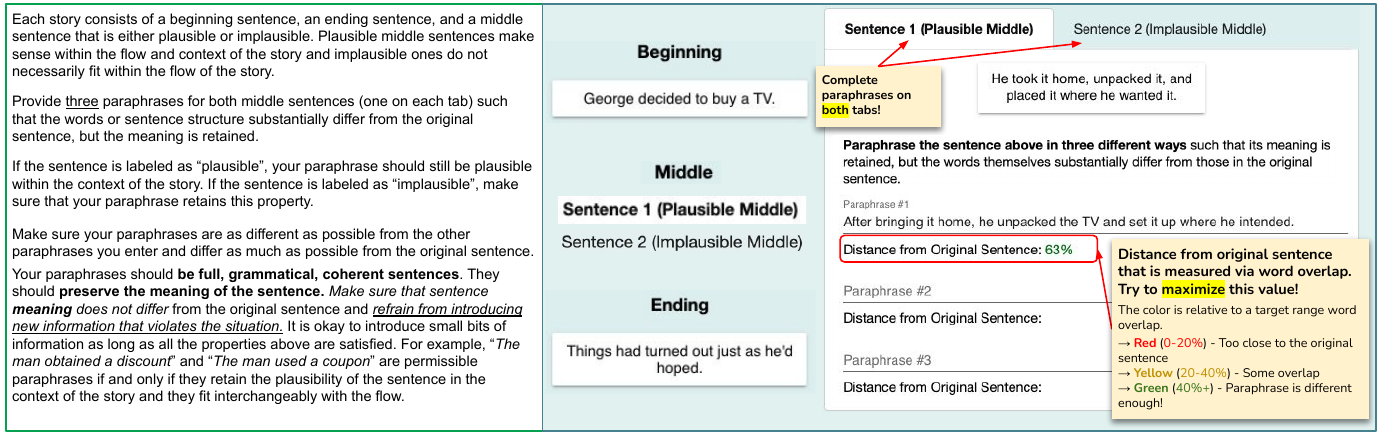}
    \end{subfigure}
    \vspace{4pt}
    \begin{subfigure}[b]{\linewidth}
        \centering
        \includegraphics[width=\textwidth]{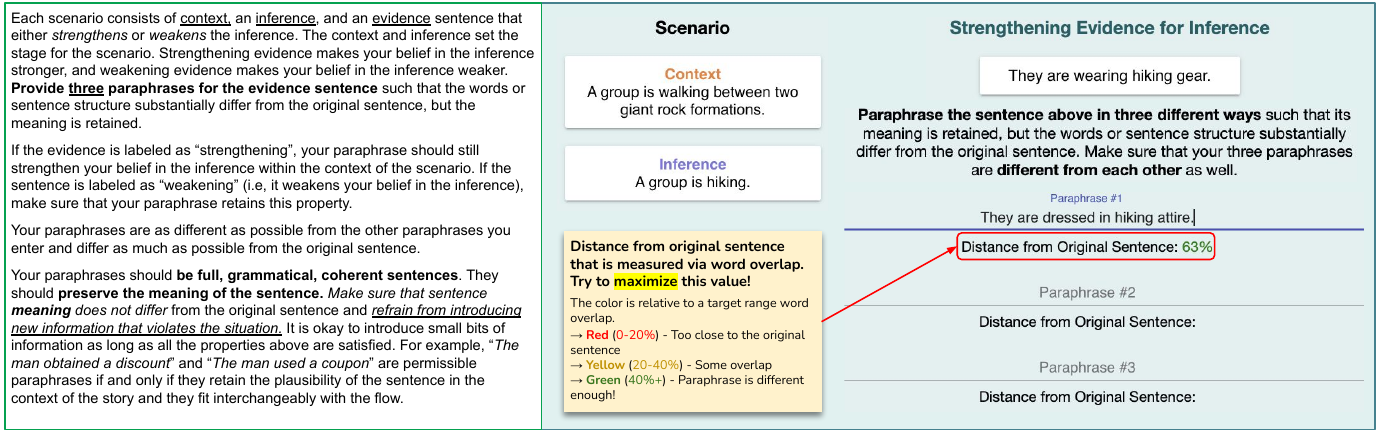}
    \end{subfigure}
    
  \caption{Annotation instructions and interface for collecting paraphrases of \anli{} (top) and \dnli{} (bottom) reasoning problems. Workers must write three label-preserving paraphrases.}
  \label{fig:interface}
  \end{minipage}
 \end{figure}

\section{Crowdsourcing \paranlu{}}
\label{appendix:crowdworking}

We collect paraphrases in \paranlu{} using Amazon Mechanical Turk.
The instructions and annotation interface shown to crowdworkers is shown in Figure~\ref{fig:interface}.

Workers provide 3 paraphrases of both plausible and implausible hypotheses for \anli{} examples and 3 paraphrases of updates for \dnli{} examples. 
We include a distance widget in our interface that computes the Jaccard similarity between the entered paraphrase and the original text to encourage lexical diversity. 
Each example was annotated by 3 workers. Workers were paid US\$12/hour on average and were required to be native English speakers with a 95\% or more HIT acceptance rate on at least 100 HITs.

\section{Paraphrastic Consistency}
\begin{figure}
     \centering
     \vspace{30em}
     \begin{subfigure}[b]{0.4\textwidth}
         \centering
         \includegraphics[scale=0.48]{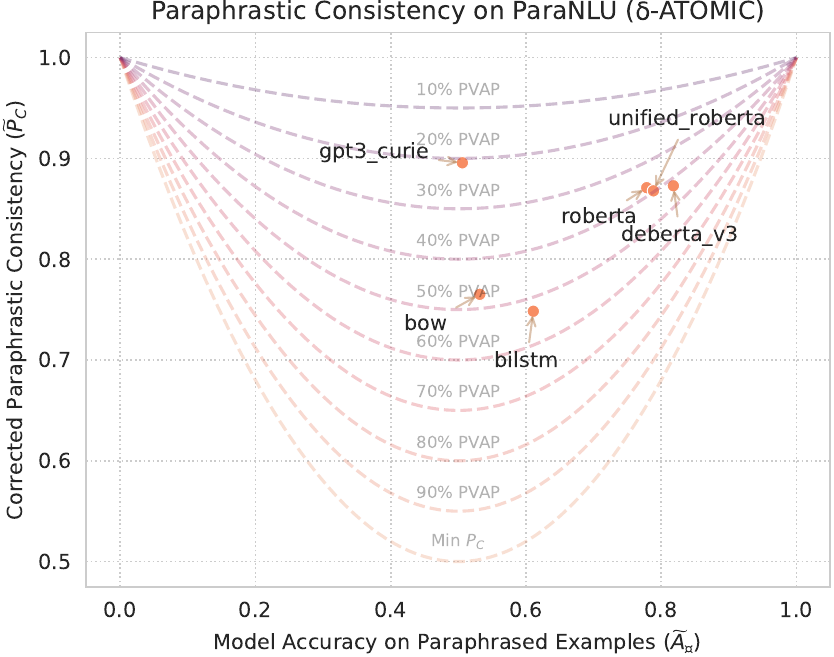}
         \vspace{-1em}
     \end{subfigure}
     \hspace{1.5em}
     \begin{subfigure}[b]{0.4\textwidth}
         \centering
         \includegraphics[scale=0.48]{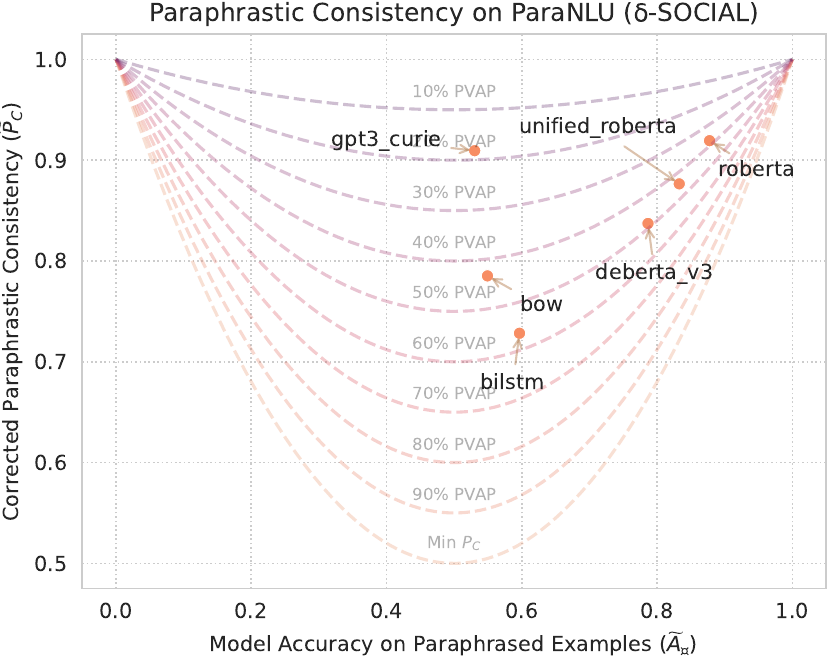}
         \vspace{-1em}
     \end{subfigure}
     \caption{Paraphrastic consistency (\pstayc{}) of different models on \atomic{} and \social{} paraphrased examples.}
     \label{fig:remaining-modeling-pstay}
     \vspace{-1.5em}
     \hfill
\end{figure}

Figure~\ref{fig:remaining-modeling-pstay} shows model accuracy plotted against corrected paraphrastic consistency of all models tested for the \atomic{} and \social{} splits of \paranlu{}.

\end{document}